\documentclass[letterpaper, 10 pt, conference]{ieeeconf}  

\IEEEoverridecommandlockouts                              

\usepackage{amsmath,amsfonts}
\usepackage[noend]{algorithmic}
\usepackage{algorithm}
\usepackage{array}
\usepackage[caption=false,font=normalsize,labelfont=sf,textfont=sf]{subfig}
\usepackage{textcomp}
\usepackage{stfloats}
\usepackage{url}
\usepackage{verbatim}
\usepackage{graphicx}
\usepackage{cite}
\usepackage{xcolor}
\usepackage{hyperref}
\usepackage{multirow}
\usepackage{booktabs}
\usepackage{pdfpages}
\usepackage{graphicx}
\usepackage{dsfont}
\usepackage{pifont}

\definecolor{mydarkblue}{rgb}{0,0.08,0.45}
\hypersetup{
    colorlinks=true,
    linkcolor=mydarkblue,
    citecolor=mydarkblue,
    filecolor=mydarkblue,
    urlcolor=mydarkblue,
    }

\begin{document}
\title{\LARGE \bf Bipedalism for Quadrupedal Robots:  Versatile Loco-Manipulation \\ through Risk-Adaptive Reinforcement Learning}
\author{Yuyou Zhang$^{1}$, Radu Corcodel$^{2}$, Ding Zhao$^{1}$
\thanks{*This work was fully supported by Mitsubishi Electric Research Labs (MERL)}
\thanks{$^{1}$Department of Mechanical Engineering, Carnegie Mellon University, Pittsburgh, PA 15213 USA, {\tt\small \{yuyouz, dingzhao\}@andrew.cmu.edu}}%
\thanks{$^{2}$Mitsubishi Electric Research Labs (MERL), Cambridge, MA 02139 USA, {\tt\small corcodel@merl.com}}
}%

\maketitle

\begin{abstract}
Loco-manipulation of quadrupedal robots has broadened robotic applications, but using legs as manipulators often compromises locomotion, while mounting arms complicates the system.
To mitigate this issue, we introduce bipedalism for quadrupedal robots, thus freeing the front legs for versatile interactions with the environment. 
We propose a risk-adaptive distributional Reinforcement Learning (RL) framework designed for quadrupedal robots walking on their hind legs, balancing worst-case conservativeness with optimal performance in this inherently unstable task. 
During training, the adaptive risk preference is dynamically adjusted based on the uncertainty of the return, measured by the coefficient of variation of the estimated return distribution. 
Extensive experiments in simulation show our method's superior performance over baselines. 
Real-world deployment on a Unitree Go2 robot further demonstrates the versatility of our policy, enabling tasks like cart pushing, obstacle probing, and payload transport, while showcasing robustness against challenging dynamics and external disturbances.

\end{abstract}


\section{Introduction}
In recent years, the field of quadrupedal robots has made remarkable progress.  In terms of locomotion,
improved capabilities of traversing various terrains and outdoor environments were developed~\cite{rudin2022learning, kumar2021rma, hoeller2024anymal, long2024hybrid, chane-sane2024soloparkour, chen2024slr, long2024learning, mitchell2024gaitor, ren2024topnav, zargarbashi2024robotkeyframing, yang2024generalized}.
Manipulation skills \cite{ha2024umi, sleiman2024guided, mendonca2024continuously, zhang2024learning, bruedigam2024a} and specialized abilities, such as ball shooting, dribbling, catching, and goalkeeping \cite{ji2022hierarchical, ji2023dribblebot, forrai2023event, huang2023creating}, have been further studied to expand the real-world applicability of quadrupedal robots.
These components enable legged robots flexible operations in unstructured environments, 
and are the building blocks of general sensorimotor skills that facilitate meaningful interactions between legged robots and their surroundings. 


One way to enable interactions with the environment is to decouple the locomotion and manipulation components by equipping the quadrupedal robots with top-mounted robotic arms \cite{ ha2024umi, sleiman2023versatile, fu2023deep} or claws \cite{lin2024locoman}.
However, adding robotic arms to quadrupedal robots significantly limits their applicability due to increased weight, energy demand, and additional spatial constraints.
Inspired by bipedalism in human evolution~\cite{hunt1994evolution}, we adopt \textit{bipedal gait} for quadrupedal robots to free up their front legs, which are typically used for locomotion in a quadrupedal gait, and repurpose them for manipulation tasks, such as pushing, environment probing, payload carrying and other tasks that require non-prehensile interactions with the surrounding environment. 


\begin{figure} [htbp]
\includegraphics[width=0.5\textwidth, page=4, trim = 0cm 0.3cm 10cm 0cm, clip]{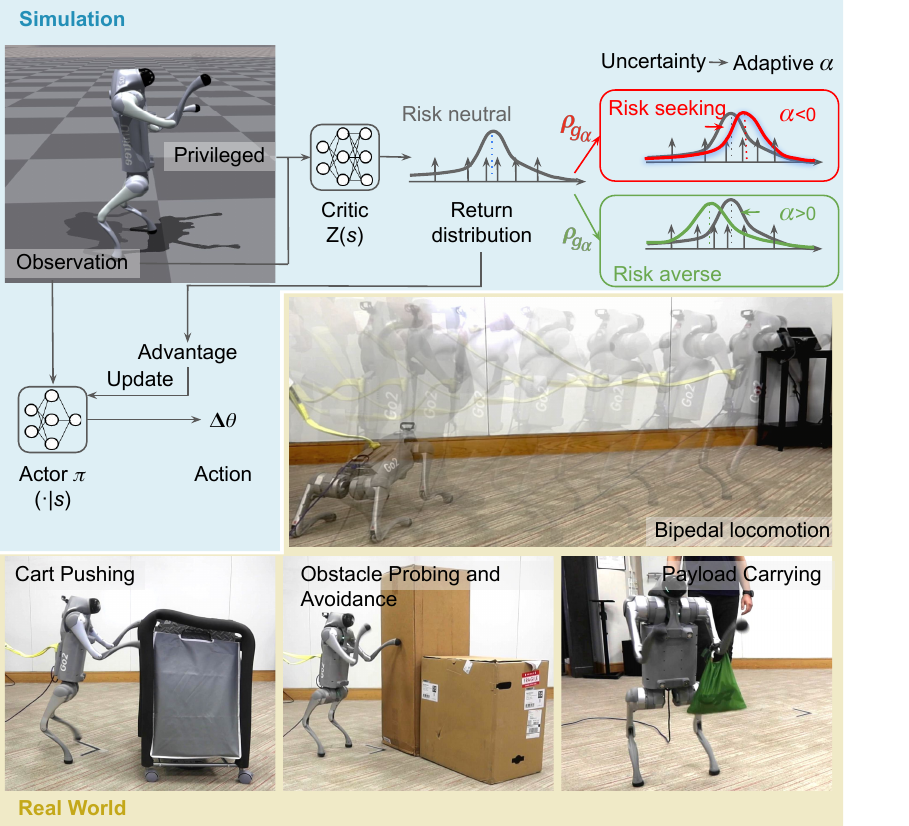}
\caption{Risk-adaptive distributional RL framework overview for bipedal locomotion on a Unitree Go2 robot. 
Using the distortion risk measures $\rho_{g_\alpha}$, the policy tends to be optimistic (\textcolor{red}{red}) in the well-explored states when the uncertainty of the return distribution is low, and vice versa (\textcolor{green}{green}). Versatile real-world applications such as cart pushing, obstacle probing, and payload carrying, are enabled by a single locomotion policy. Demonstrations are available in our supplementary video.
}
\label{fig: overview}
\vspace{-15pt}
\end{figure}

Bipedal locomotion differs from quadrupedal locomotion in its inherent instability~\cite{grizzle2014models} due to the narrower base of support. 
These challenges can be exacerbated in the real world when dynamics and disturbances are unknown~\cite{akella2024risk}.
RL is commonly used to learn a control policy with complex dynamics. RL in standard form is usually risk-neutral and maximizes the expected accumulated return~\cite{schulman2017proximal, Brendan2023epistemic}.
The sim-to-real gap and unexpected perturbations during deployment can destabilize a risk-neutral policy, which focuses only on expected return,
especially when worst-case returns are underrepresented in the return distribution~\cite{akella2024risk}.
It is thus crucial to adopt a risk-aware approach to consider these worst-case scenarios, particularly when the unpredictability of real-world deployment poses significant risks.



In this work, we propose a risk-adaptive distributional RL framework, as shown in Figure \ref{fig: overview}, to learn a robust policy for inherently unstable bipedal locomotion. 
Specifically, during training, we adapt the risk preference dynamically based on the uncertainty of the return, measured by the coefficient of variation of the estimated return distribution, instead of pre-specifying the risk level for policy learning. 
Extensive simulation experiments demonstrate the superior performance of our method compared to baseline approaches.
In real-world deployment, we showcase loco-manipulation including cart pushing, contact-aware obstacle probing, and payload carrying, highlighting the versatility and robustness of the bipedal locomotion policy.
In summary, the main contributions of this work are:
\begin{itemize} 
    \item We introduce a risk-adaptive RL framework for the robust bipedal locomotion of quadrupedal robots. 
    \item We propose a novel uncertainty metric based on the return distribution to adaptively choose the risk level.
    \item We demonstrate robust real-world applications with bipedal locomotion under external force, and highlight three representative tasks including cart pushing, contact-aware obstacle probing, and payload carrying.
\end{itemize}


\section{Related Work}
\subsection{Quadrupedal robot locomotion and manipulation}
Previous work either equips the legged robot with a mounted robotic arm~\cite{bellicoso2019alma, sleiman2021unified, sleiman2023versatile, fu2023deep, ha2024umi, sleiman2024guided, mendonca2024continuously, zhang2024learning, liu2024visual, yokoyama2023asc, portela2024learning, ma2022combining, zhang2024gamma} or use one leg as manipulator~\cite{he2024learning, huang2024hilma, arm2024pedipulate, ji2022hierarchical, ji2023dribblebot}.
With mounted arms, mobile manipulation requires coordination between the robot arm and the legged robot.
Instead of decoupling the manipulation and locomotion controllers like in~\cite{bellicoso2019alma, ma2022combining},
more recent work seeks to build manipulation-centric whole-body controllers to allow better coordination: ~\cite{sleiman2021unified, sleiman2023versatile} formulate a unified whole-body MPC framework, 
~\cite{fu2023deep} uses RL to train whole-body control policy for end-effector tracking,~\cite{liu2024visual} trains RL policy with vision input,~\cite{ha2024umi} uses diffusion policy to learn from human demonstration,~\cite{portela2024learning} learns a whole-body force control policy to enable compliance and force application.
However, adding a mounted arm to a quadrupedal robot increases the load and energy requirements, and unnecessarily adds system complexity.

Loco-manipulation repurposes the robot's legs for manipulation without changing its embodiment. 
However, increased manipulation ability comes at the cost of compromised locomotion ability since the robot's legs are primarily designed for locomotion.
For example, most loco-manipulations have the robots stand still on three legs and use one leg as the manipulator~\cite{arm2024pedipulate, he2024visual, lin2024locoman, he2024learning, ouyang2024long, cheng2023legs}. 
Inspired by how bipedalism played an important role in human evolution by freeing the hands for manipulation~\cite{hunt1994evolution}, we adopt a bipedal gait to enable flexible loco-manipulation such as bi-manual pushing, contact-aware obstacle probing, and payload carrying.
Compared to previous work on bipedal locomotion of quadrupedal robots~\cite{li2023learning, su2024leveraging, long2024learning}, our work further explores meaningful interactions between quadrupedal robots and their surrounding environment, enabled by our robust bipedal locomotion policy.



\subsection{Risk-aware RL for Robot}
Risk awareness is essential for the successful real-world deployment of autonomous robots, such as drones~\cite{liu2023adaptive} and quadrupedal robots~\cite{schneider2024learning, li2024learning, shi2024robust, fan2021learning}. Distributional RL~\cite{bellemare2017distributional, dabney2018distributional, dabney2018implicit} models the distribution of returns explicitly, rather than estimating the value function as the expected return, making it widely applicable in risk-sensitive RL~\cite{greenberg2022efficient, clements2019estimating}.

Several works have used distributional RL to improve the robustness of quadrupedal robot real-world performance. 
Li et al. \cite{li2024learning} introduce a distribution ensemble actor-critic approach and demonstrate improved performance in domain randomization settings. 
However, the approach is not risk-aware.
Shi et al. ~\cite{shi2024robust} takes a risk-adaptive perspective but can only switch between a risk-neutral (CVaR$_1$) policy and a risk-averse (CVaR$_{0.5}$) policy, resembling switch-mode control. Also, risk-averse learning ~\cite{shi2024robust} can ignore high-return strategies ~\cite{greenberg2022efficient}.
Schneider et al. \cite{schneider2024learning} propose a risk-aware Distributional PPO and demonstrate risk-aware locomotion behavior conditioned on a manually specified risk level at deployment, which relies on human prior knowledge. 
Different from previous work, our method enables risk adaptiveness during training, allowing the actor policy to internalize risk preference selection automatically. During training, the value estimation becomes conservative in high-uncertainty situations and autonomously shifts toward optimism when the return distribution exhibits low variance. This results in a policy that is both robust and high-performing, without requiring manual risk tuning at deployment.



\section{Preliminary}
\textbf{Partially Observable Markov Decision Process.}
We formulate bipedal locomotion learning as a Partially Observable Markov Decision Process (POMDP) defined by \((S, \mathcal{A}, \mathcal{T}, R, \Omega, O, \gamma)\), where \(S\) represents the state space, \(\mathcal{A}\) the action space, \(\mathcal{T} : S \times \mathcal{A} \mapsto S\) is the transition function, \(R : S \times \mathcal{A} \mapsto \mathbb{R}\) is the reward function, $\Omega$ is the set of observations,  $O$ is the observation function, \(\gamma\) is the discount factor. The objective is to train a policy \(\pi^*\) which maximizes the discounted cumulative reward
$
\pi^* = \arg\max_{\pi} \mathbb{E}_{s_0 \sim \rho_0, a_t \sim \pi(\cdot | s_t)} \left[ \sum_{t \geq 0} \gamma^t r(s_t, a_t) \right]$.

\textbf{Distributional RL.}
Distributional RL~\cite{bellemare2017distributional, dabney2018distributional,dabney2018implicit} learns the value distribution, instead of the expected return as a value function.
With policy $\pi$, the return is a random variable $Z^\pi$ that represents the cumulated discounted rewards along one trajectory, $Z^\pi = \sum_{t=0}^{\infty} \gamma^t R_t$.
The value function for many standard RL algorithms is, 
$
V^\pi(x) = \mathbb{E}[Z^\pi(x)] 
$
while distributional RL explicitly parameterizes the return distribution with quantile functions ~\cite{dabney2018distributional,dabney2018implicit} or discrete distribution ~\cite{bellemare2017distributional}.
We adopt a similar parameterization as in QR-DQN~\cite{dabney2018distributional}, where the return distribution is approximated by estimating the quantiles $\tau_1, \cdots, \tau_N, \tau_i= i/N, i=1, \cdots, N$, with the parametric model $\theta$,
%
\begin{equation}
    Z_{\theta}(x) := \frac{1}{N} \sum_{i=1}^{N} \delta_{\theta_i(x)},
    \label{eq: preliminary}
\end{equation}
where $\theta_i(x)$ is the $i^{th}$ quantile of the return distribution $Z^\pi(x)$, and $\delta_{\theta_i(x)}$ denotes a Dirac function at $\theta_i(x)$.

\section{Method}

We first introduce the problem setting of the bipedal locomotion task in section \ref{method: environment}. Due to the robustness requirement of balance in bipedal locomotion, we adopt distributional RL shown in section \ref{method: distributional}. In section \ref{method: risk_adaptive}, we incorporate adaptive risk measures to balance safety in worst-case scenarios with optimal task performance.
The framework overview of our proposed method is shown in Figure \ref{fig: overview}. 

\subsection{Control Policy with RL}
\label{method: environment}

\textbf{Observation and Action Space.} 
The bipedal locomotion policy receives observations which include the proprioceptive information, locomotion command, and the last action $a_{t-1} \in \mathbb{R}^{12}$.  
Proprioceptive information includes joint position $\theta_t \in \mathbb{R}^{12}$ and joint velocity $\dot{\theta}_t \in \mathbb{R}^{12}$ provided by the joint encoders and projected gravity in the robot frame $g_t\in \mathbb{R}^{3}$ from the IMU. 
The command $\mathbf{c}_t = [v_x^c, v_y^c, \omega_{\text{yaw}}^c, z^c, f^c]$ includes the velocity command specifying the linear velocities in the longitudinal and lateral directions, angular velocity around the vertical axis, base height, and stepping frequency.
Privileged observation for the critic network at the training stage includes extra information only available in simulation such as joint friction and restitution coefficient.
The dimension of the action space $\mathcal{A}$ is $12$, which equals the number of actuators. 
The predictions of the policy, $\Delta \theta_t \in \mathbb{R}^{12}$, are the joint angles relative to the nominal quadrupedal standing position. 

\textbf{Reward Functions.} 
\label{sec: reward}
Reward functions consist of task-specific rewards for bipedal locomotion and auxiliary rewards adapted from~\cite{margolis2023walk} to optimize foot contact, action smoothness, energy consumption, joint position, etc.
Task-specific reward functions are listed in Table \ref{tab: reward}. 
In addition to 
\textit{Base Height} which
encourages maintaining base height $z^c$ and \textit{Base Pitch} to promote an upright position, we also use \textit{Upright Balance} to penalize velocity along the $ z $-axis and changes in pitch angle $ \dot{p} $ when the robot is upright. 
Velocity tracking rewards include \textit{Linear Tracking} and \textit{Angular Tracking}, where $\sigma$ and $\sigma_{\rm yaw}$ are scaling factors.

\begin{figure} [htbp]
\includegraphics[width=0.48\textwidth, page=2, trim = 3cm 5.2cm 3cm 2cm, clip]{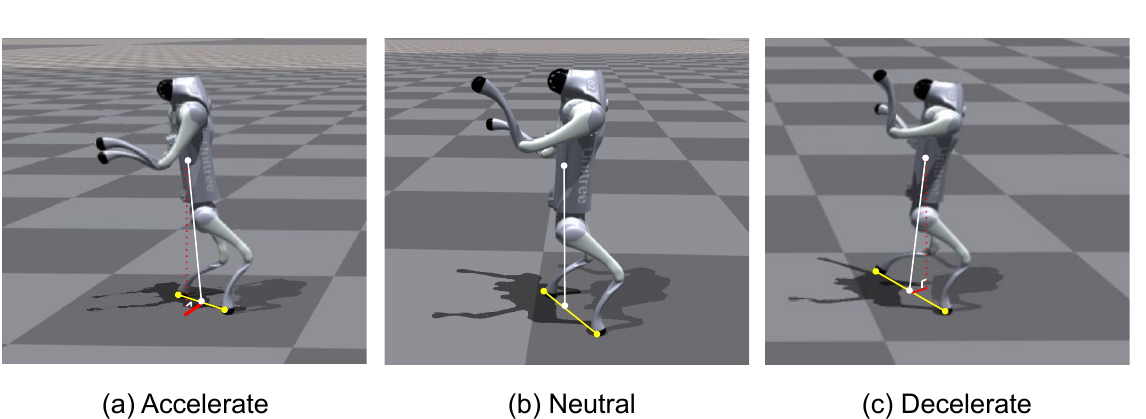}
\centering
\caption{The robot accelerates (a), stays neutral (b), and decelerates (c) by shifting its center of mass (CoM) ahead of, aligned with, or behind the support polygon.
}
\label{fig: CoM}
\end{figure}

Apart from direct tracking reward,  we design a \textit{Support Polygon} reward to track the relative position between the base center of mass (CoM) and the support polygon.
When the CoM moves ahead of the support polygon, the robot accelerates. Conversely, when the CoM lags behind, the robot decelerates.
This mechanism enables continuous balance control while tracking the desired velocity, as shown in Fig. \ref{fig: CoM}.
We characterize the relative position between the robot CoM and support polygon by $\arctan(\Delta x_b/\Delta z_b)$.
$\Delta x_b$ and $\Delta z_b$ are relative positions of the average of two rear feet in the body frame along the x-axis and z-axis. 
$\arctan(\Delta x_b/\Delta z_b)$ is expected to be positive when the robot needs to accelerate and negative when decelerating.
This angle is essentially different than the pitch angle and only degenerates to the pitch angle when the model is simplified to a single inverted pendulum.
The total reward is a positive linear function of the task reward, $r_{+} \times e^{c \times r_{-}}$, where $r_{+}$ is the sum of positive reward terms and $r_{-}$ is the sum of
negative reward terms, $c$ is coefficient set to $0.02$. 
\begin{table}[t]
\caption{Task reward functions}
\vspace{-2ex}
\scalebox{0.84}{
\begin{tabular}{p{2.0cm} p{7.2cm}}
\toprule
\textit{Base Height}                         & $-(z - z^{\text{c}})^2$
\\ \midrule
\textit{Base Pitch}      &  $-\cos(p^c - p) $
\\ \midrule
\textit{Upright Balance }          & $ \exp(-v_z^2/\sigma) +  \exp(-\dot{p}^2/\sigma_{\rm yaw}) \; \text{if is upright, else } 0 $  
\\ \midrule
\textit{Linear Tracking}           & $ \exp(-|v_{xy}-v_{xy}^c|^2/\sigma ) \; \text{if is upright, else } 0 $ 
\\ \midrule
\textit{Angular Tracking}           & $ \exp(-|w_{\rm yaw}- w_{\rm yaw}^c|^2\sigma_{\rm yaw}) \; \text{if is upright, else } 0 $ 
\\ \midrule
\textit{Support Polygon}  & $
-|v_x^c|^2\left(\frac{\pi}{2}-|\arctan(\frac{\Delta x_b}{\Delta z_b})|\right)^2 \; \text{if} \arctan(\frac{\Delta x_b}{\Delta z_b}) v_x^c < 0 
$
\\ \bottomrule                    
\end{tabular}}
\label{tab: reward}
\vspace{-10pt}
\end{table}

\subsection{Risk-aware Distributional PPO}
\label{method: distributional}
In Distributional PPO, the actor network remains the same as in standard PPO~\cite{schulman2017proximal} and is trained to optimize the clipped objective,
\begin{equation}
    \mathcal{L}(\phi) = \mathbb{E}_t \left[ \text{min}( \eta_t(\phi)\hat{A}_t, \text{clip}( \eta_t(\phi), 1 - \epsilon, 1 + \epsilon) \hat{A}_t)\right],
\end{equation}
where $\displaystyle{\eta_t(\phi) = \frac{\pi_{\phi}(a_t|o_t)}{\pi_{\phi_{\rm old}}(a_t|o_tt)}}$, is the probability ratio between the new policy $\pi_{\phi}$, and old policy $\pi_{\phi_{\rm old}}$, $\hat{A}_t$ is the estimated advantage computed using the Generalizable Advantage Estimation (GAE)~\cite{schulman2015high}.
The critic network differs from standard PPO by predicting the quantiles of the value distribution 
as in Equation \ref{eq: preliminary}, instead of just the expected value. 
$\theta_i(x) = F_Z^{-1}(\tau_i)$ for $\tau_i=i/N, i=1, \cdots, N$ are $N-$quantiles of the return distribution $Z^\pi(x)$, 
and $F_Z^{-1}$ denotes the inverse CDF of the return distribution $Z^\pi(x)$.
For a critic network that approximate the return distribution by predicting $\theta_1(x), \cdots, \theta_N(x)$, the objective is to minimize the following quantile loss, which effectively minimizes the 1-Wasserstein distance between the empirical distribution $\hat{Z}^\pi(x)$ and the parameterized quantile distribution $ Z_\theta^\pi(x)$:
\begin{equation}
    \mathcal{L}(\theta) = \mathbb{E}_t\left[\frac{1}{N}\sum_{i=1}^N (\tau - \mathds{1}_{z_t - \theta^i_t<0}) (z_t - \theta^i_t) \right],
\end{equation}
where $z_i\sim Z^\pi(x_t)$, and $\theta^i_t= \theta_i(x_t)$.

The value function derived from the return distribution is used to estimate the advantage $\hat{A}_t$ for the actor-network update. Given the return distribution predicted by the critic network, the risk-neutral value function is calculated as,
\begin{equation}
    V(x) = \sum_{i=1}^{N}(\tau_i - \tau_{i-1}) \theta_i(x) = \sum_{i=1}^{N}\frac{1}{N}\theta_i(x).
\end{equation}
To learn a risk-aware policy, the distortion risk measures $\rho_{g_\alpha}$ associated with the distortion function $g_\alpha$ is applied to the return distribution. Conditional Value at Risk (CVaR) is commonly applied to have risk-averse behaviors since only the left tail is quantified. 
We apply Wang's metric ~\cite{wang2000class} so that the risk preference can be adjusted from averse ($\alpha>0$) to  seeking ($\alpha<0$) as in equation \ref{eqn: wang},
\begin{equation}
    g_{\alpha}^{\text{Wang}}(\tau)  = \Phi(\Phi^{-1}(\tau) + \alpha).
\label{eqn: wang}
\end{equation}
Then the new value function is given by the distorted return distribution as in Equation \ref{eq: risk_measure}.
\begin{equation}
    V_\alpha(x) = \rho_{g_\alpha}(Z_\theta^\pi(x))= \sum_{i=1}^{N} \left( g_{\alpha}(\tau_i) - g_{\alpha}(\tau_{i-1}) \right) \theta_i(x).
    \label{eq: risk_measure}
\end{equation}
When $\alpha>0$, the calculation of the value function is conservative, by assigning more weight to worst-case left tails, while $\alpha<0$ makes the calculation optimistic by assigning more weight to higher returns.

\subsection{Uncertainty Modeling and Adaptive Risk Level}
\label{method: risk_adaptive}
We assume the transition is deterministic in this POMDP, then the aleatory uncertainty comes from partial observation of the state and domain randomization, while the epistemic uncertainty arises from the lack of environment knowledge to make informed predictions.
Optimism in the face of epistemic uncertainty encourages exploration, allowing for the collection of more informative data to maximize long-term returns~\cite{Brendan2023epistemic}. However, conservativeness is essential to ensure worst-case performance when aleatory uncertainty is present.
We model the uncertainty as the uncertainty of the parameterized distribution $Z_\theta^\pi(x)$ predicted by the critic network, represented by the Coefficient of Variation (CV),
\begin{equation}
    CV_{Z_\theta^\pi(x)} = \frac{\sqrt{{\rm Var}(Z_\theta^\pi(x))}}{\mathbb{E}Z_\theta^\pi(x)} = \frac{\sigma}{\mu}.
\end{equation}
This normalized measure allows for the comparison of dispersion across different return distributions, even if the means are drastically different from each other.
It is particularly advantageous over non-normalized measures, such as right truncated variance (RTV) used in~\cite{liu2023adaptive} and interquartile range (IQR) used in~\cite{shi2024robust}, as the mean of the return distribution tends to increase significantly during training due to increased rewards.

We formulate the parameter $\alpha$ of the distortion function $g_{\alpha}^{\text{Wang}}(\tau)$ as a function of the modeled uncertainty $CV_{Z_\theta^\pi(x)}$ and training steps $t$, 
\begin{equation}
    \alpha_t = (\alpha_0 - \alpha_T) e^{-\frac{t/T}{CV_t}} +\alpha_T  , 
    \label{eq: risk_adaptive}
\end{equation}
where $T$ is the total training steps. And $CV_t$ is  the average over batch of $CV_{Z_\theta^\pi(x)}$ at step $t$. 
With $\alpha_0 - \alpha_T>0$, the policy begins conservatively and becomes increasingly optimistic as training progresses. The dependence of $\alpha_t$ on $CV_t$ directs the policy to be more conservative when uncertainty is high. The time-dependent coefficient $\frac{t}{T}$ allows $\alpha_t$ to be progressively more influenced by the uncertainty $CV_t$ as training advances. 

\section{Experiments}
We evaluate our method across several experiments. In Section \ref{sec: curve}, we demonstrate our method has higher training rewards compared to baseline methods and ablations. In Section \ref{sec: error}, we evaluate velocity tracking error across in-distribution and out-of-distribution target velocities, as well as under external forces. In Section \ref{sec: vis}, we analyze uncertainty and risk modeling to highlight the importance of risk-adaptive learning. Finally, in Section \ref{sec: real}, we deploy our bipedal locomotion policy on the Unitree Go2 robot, showing that a single policy enables robust bipedal locomotion and versatile loco-manipulation capabilities.

\textbf{Simulation setup}
We use Isaac Gym~\cite{makoviychuk2021isaac} to train the bipedal locomotion policy based on the open-source framework in \cite{margolis2023walk}. 
The target velocity range is $[-0.8, 0.8]$ m/s for $v_x^c$ and is $[-0.4, 0.4]$ m/s for $v_y^c$, and $[-1, 1]$ rad for $\omega_{\rm yaw}$. 
The critic network and actor network both have hidden dimensions $[512, 256, 128]$. 
The output layer size of the critic network is 64, to predict $N=64$ quantiles of the return distribution. 

We initialize our risk-adaptive DPPO with neutral initial risk $\alpha_0 = 0$ to maintain stability and avoid early catastrophic failures during training. However, remaining too conservative may hinder exploration and limit performance improvements. Therefore, we gradually shift to a more optimistic risk preference $\alpha_T = -0.2$, encouraging the agent to explore high-return strategies and accelerate learning.
We select these parameters to avoid over-conservativeness early on, which may prevent discovering successful quadrupedal-to-bipedal transitions, and to limit excessive optimism later, which could destabilize training. 
We train 4000 agents in parallel for 20k
iterations on an NVIDIA RTX 4090 GPU, which takes approximately 5 hours. 
We compare our method to baselines in Table \ref{tab: baseline}. All methods share the same hyperparameters if applicable.

\textbf{Hardware setup}
We use the Unitree Go2 robot for real-world experiments.  The computations are performed on a host computer. The policy runs at 50Hz and the robot receives the joint position command from the host computer. 
Target joint angles were tracked using a PD controller with gains set to $K_p = 25$ and $K_d = 0.6$.

\subsection{Training Performance}
\label{sec: curve}

\begin{figure} [htbp]
\includegraphics[width=0.49\textwidth, page=1, trim = 0.8cm 3cm 2.9cm 4.6cm, clip]{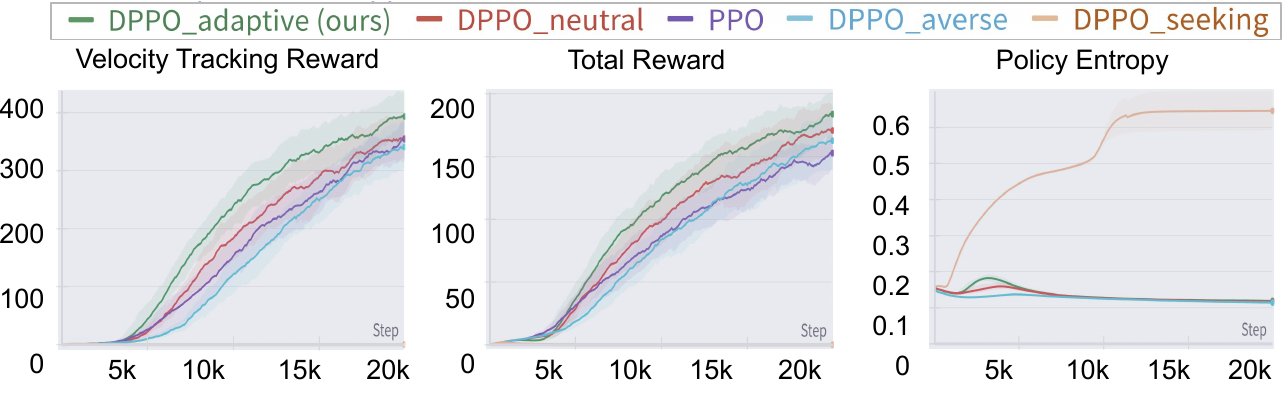}
\vspace{-4.5ex}
\caption{Learning curves of proposed method (\textit{DPPO\_adaptive}) against baselines listed in Table \ref{tab: baseline}. The rewards are averaged over three seeds, and the shaded region represents the standard error.
}
\label{fig: baseline}
\end{figure}

\textbf{Baseline comparison}
Figure \ref{fig: baseline} shows the comparison between proposed risk-adaptive distributional PPO (DPPO) and baselines in Table \ref{tab: baseline}. 
Parameter $\alpha$ in the distortion function  $g_{\alpha}^{\text{Wang}}(\tau)$ for 
\textit{DPPO\_averse} is 0.2, and for \textit{DPPO\_seeking} is $-0.2$, which equals to the final risk $\alpha_T$ for \textit{DPPO\_adaptive}.

\begin{table}[t]
\caption{Proposed Method and Baselines Settings}
\vspace{-2ex}
\centering
\scalebox{0.84}{
\begin{tabular}{cccc}
\toprule
\textbf{Method}         & \textbf{Distributional} & \textbf{Fixed Risk} & \textbf{Adaptive Risk} \\
\midrule
\textit{DPPO\_adaptive} (ours) & \checkmark  & \ding{55}   & \checkmark \\
\midrule
\textit{DPPO\_neutral}~\cite{schneider2024learning}         & \checkmark  & $\alpha = 0$ & \ding{55} \\
\midrule
\textit{DPPO\_averse}~\cite{schneider2024learning}         & \checkmark  & $\alpha > 0$ & \ding{55} \\
\midrule
\textit{DPPO\_seeking }~\cite{schneider2024learning}        & \checkmark  & $\alpha <0$ & \ding{55} \\
\midrule
\textit{PPO}~\cite{schulman2017proximal}                   & \ding{55}   & \ding{55} & \ding{55} \\
\bottomrule
\end{tabular}
}
\vspace{-4ex}
\label{tab: baseline}
\end{table}

Our method consistently outperforms the baselines in both velocity tracking and total reward.
Risk-neutral DPPO and PPO perform similarly and both achieve a lower total reward compared to DPPO with adaptive risk.
Risk-seeking DPPO fails to learn a locomotion policy, resulting in near-zero rewards after 20k steps, and diverging policy entropy, indicating that constant risk-seeking can lead to catastrophic failures.
In contrast, risk-adaptive DPPO exhibits higher policy entropy during training compared to the risk-neutral baselines (Risk-neutral DPPO and PPO), suggesting that our method encourages exploration of diverse actions, while the risk-neutral policies are less exploratory.
Risk-averse DPPO achieves the lowest velocity tracking reward but ranks second-to-last in total reward, due to its risk-averse strategy, which minimizes the accumulation of negative penalties that contribute to the total reward.

\begin{figure} [htbp]
\includegraphics[width=0.49\textwidth, page=5, trim = 0.6cm 3cm 2.6cm 4.5cm, clip]{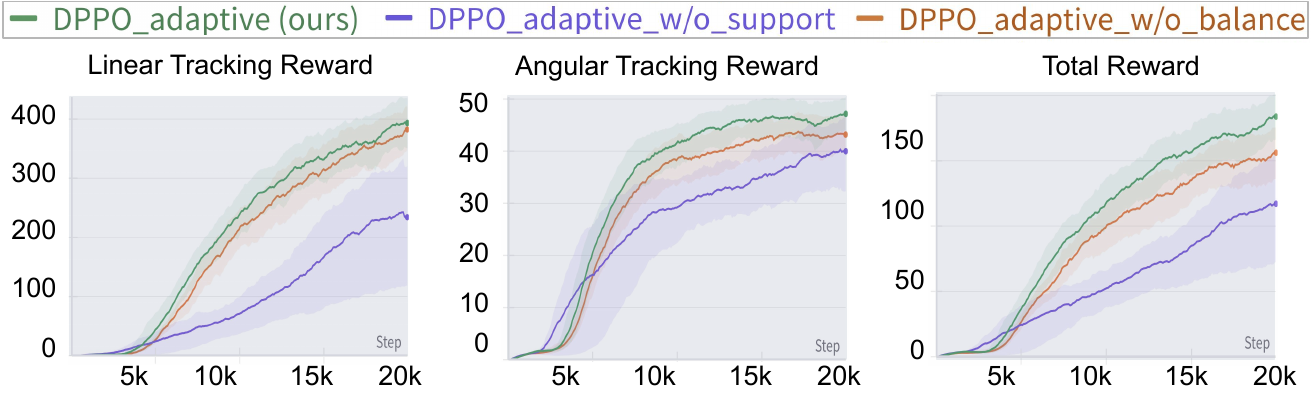}
\vspace{-4.5ex}
\caption{Learning curves of our method and reward function ablations. The rewards are averaged over three seeds, and the shaded region represents the standard error.
}
\label{fig: ablation}
\vspace{-5pt}
\end{figure}

\textbf{Ablations of reward functions}
To assess the impact of proposed reward functions in Table \ref{tab: reward}, we compare our method to \textit{DPPO\_adaptive\_w/o\_support} and \textit{DPPO\_adaptive\_w/o\_balance}, where, in each case, one of the task reward functions is removed. As shown in Figure \ref{fig: ablation},  the absence of the \textit{Support Polygon} reward function leads to a significant drop in linear velocity tracking performance, and the \textit{Upright Balance} reward function enhances overall bipedal locomotion performance.

\subsection{Tracking Error Evaluation}
\label{sec: error}
We evaluate the learned policy based on success rate and velocity tracking error. The evaluation is averaged across 4000 environments, each with an episode length of 1000 steps. An episode is considered successful if it does not terminate early due to the robot crashing. The tracking error is calculated as the Root mean square error (RMSE) across all evaluation environments and episode steps.

\begin{figure}[t] 
    \centering
    \begin{minipage}{0.24\textwidth}
        \centering
        \includegraphics[width=\linewidth]{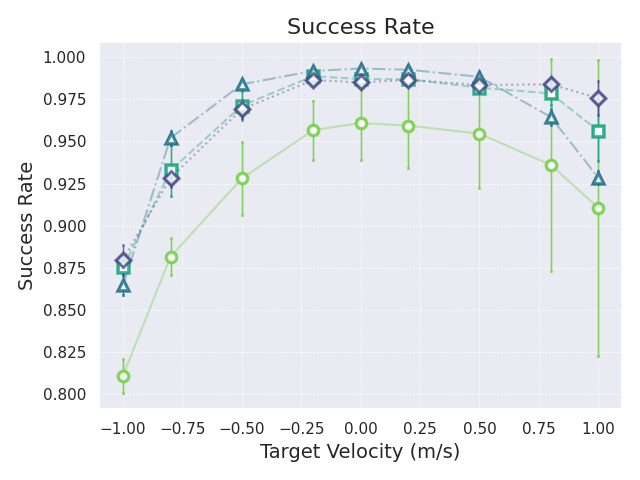}
        (a) 
    \end{minipage}\hfill
    \begin{minipage}{0.24\textwidth}
        \centering
        \includegraphics[width=\linewidth]{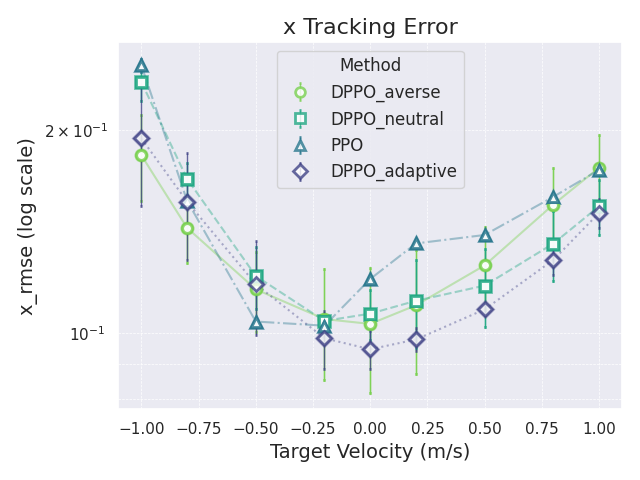}
        (b) 
    \end{minipage}\hfill
    
    \caption{Success Rate (a) and X Tracking Error (b) across target velocities ranging from -1.0 m/s to 1.0 m/s. 
    Comparison of \textit{DPPO\_adaptive} with three baseline methods.
    }
    \label{fig: evaluate}
\end{figure}

\textbf{Across Varying Target Velocities}
With the training target velocity $v_x$  sampled in the range of $[-0.8, 0.8]$ m/s, we use in-distribution velocities of $[\pm0.8, \pm0.5, \pm0.2, 0.0]$ m/s and out-of-distribution (OOD) velocities of $\pm 1.0$ m/s as the evaluation target velocities.
We show that risk-adaptive DPPO outperforms baselines with the highest success rate and lowest tracking error in Table \ref{tab: average}.
More specifically, risk-adaptive DPPO achieves the lowest tracking error for target velocities of -0.25 m/s or higher in Figure \ref{fig: evaluate}. 
For target velocities below -0.25 m/s, PPO and risk-averse DPPO perform better, suggesting that backward velocity tracking may require a more conservative policy.
This is consistently indicated by the generally lower success rate for backward tracking compared to forward velocity tracking. 
Despite this, we did not fully explore the potential of our method by tuning the initial and final risk levels, as forward velocity tracking is more common in real-world deployments.

\begin{table}[htbp]
\centering
\caption{Average Velocity Tracking Error and Success Rate across Different Target Velocity}
\vspace{-2ex}
\label{tab: average}
\resizebox{0.4\textwidth}{!}{%
\begin{tabular}{@{}lccc@{}}
\toprule
\multirow{1}{*}{Method}
 & \multicolumn{1}{c}{Success Rate$\uparrow$} 
 & \multicolumn{1}{c}{x\_RMSE $\downarrow$} 
 & \multicolumn{1}{c}{y\_RMSE $\downarrow$}

 \\\midrule
\multirow{1}{*}{\textbf{DPPO\_adaptive (ours)}} 
& $\textbf{0.964}\pm0.0044$  & $\textbf{0.128}\pm0.0100$  & $\textbf{0.072}\pm0.0002$
\\ 
\midrule

\multirow{1}{*}{DPPO\_averse} ~\cite{schneider2024learning}  
& $0.922\pm0.0142$  & $0.135\pm0.0088$  & $0.080\pm0.0028$      
\\ 
\midrule

\multirow{1}{*}{DPPO\_neutral} ~\cite{schneider2024learning}  
& $0.962\pm0.0006$  & $0.139\pm0.0035$  & $0.074\pm0.0010$  
\\ 
\midrule

\multirow{1}{*}{PPO}
& $0.962\pm0.0011$  & $0.149\pm0.0003$  & $0.077\pm0.0002$  
\\ 
\midrule

\bottomrule
\end{tabular}%
}
\end{table}

Even though backward target velocity prefers a more conservative policy, we show that risk-adaptive DPPO with a risk-seeking tendency outperforms neutral and risk-averse baselines, showing significant generalizability when evaluated with OOD velocity command $\pm 1.0$ m/s, including negative backward velocity $-1$ m/s, as shown in Table \ref{tab: robustness_ood_vel}.

\begin{table}[t]
\centering
\caption{Out of Distribution Target Velocity Tracking Error}
\vspace{-2ex}
\label{tab: robustness_ood_vel}
\resizebox{0.49\textwidth}{!}{%
\begin{tabular}{@{}lcccccc@{}}
\toprule
\multirow{2}{*}{Method} & \multicolumn{3}{c|}{1m/s}& \multicolumn{3}{c}{-1m/s} \\ 
\cmidrule(l){2-7} 
 & \multicolumn{1}{c}{Success Rate $\uparrow$} 
 & \multicolumn{1}{c}{x\_RMSE $\downarrow$} 
 & \multicolumn{1}{c}{y\_RMSE $\downarrow$} 

 & \multicolumn{1}{c}{Success Rate$\uparrow$} 
 & \multicolumn{1}{c}{x\_RMSE $\downarrow$} 
 & \multicolumn{1}{c}{y\_RMSE $\downarrow$}
 \\\midrule
\multirow{1}{*}{\textbf{DPPO\_adaptive}} 
& $\textbf{0.976}\pm0.009$  & $\textbf{0.151}\pm0.007$  & $\textbf{0.072}\pm0.001$
& $\textbf{0.880}\pm0.009$  & $0.194\pm0.039$           & $\textbf{0.087}\pm0.007$
\\ 
\midrule

\multirow{1}{*}{DPPO\_averse} ~\cite{schneider2024learning} 
& $0.911\pm0.088$           & $0.175\pm0.021$  & $0.086\pm0.010$
& $0.819\pm0.010$           & $\textbf{0.183}\pm0.026$  & $0.094\pm0.009$
\\ 
\midrule

\multirow{1}{*}{DPPO\_neutral} ~\cite{schneider2024learning} 
& $0.956\pm0.018$           & $0.154\pm0.014$  & $0.083\pm0.005$ 
& $0.875\pm0.003$           & $0.235\pm0.014$  & $0.088\pm0.008$
\\ 
\midrule

\multirow{1}{*}{PPO} 
& $0.929\pm0.004$           & $0.174\pm0.005$  & $0.097\pm0.004$
& $0.864\pm0.006$           & $0.249\pm0.005$  & $0.098\pm0.001$
\\ 
\midrule

\bottomrule
\end{tabular}%
}
\end{table}

\textbf{Under external force}
We evaluate the performance under external force to further assess the robustness of our proposed method, as shown in Table \ref{tab: robustness_force}. A 10N external force was applied downward on each of the robot's forearms with a velocity command of 1m/s. Our method achieves the highest success rate, nearly doubling the second-best, and also exhibits the smallest drop in success rate compared to conditions without external force. Although risk-averse DPPO shows a slightly smaller X tracking error, this doesn't indicate better performance, as its success rate is only half that of our method. The lower error is likely induced by early-terminated episodes, which could have exhibited significantly higher tracking errors if they had not failed. 
Our method enables robust adaptation to disturbances while maintaining high performance by dynamically adjusting risk based on value function uncertainty during training, which is further explained in Section~\ref{sec: vis}.
\begin{table}[t]
\centering
\caption{Velocity Tracking Error under External Force}
\vspace{-2ex}
\label{tab: robustness_force}
\resizebox{0.4\textwidth}{!}{%
\begin{tabular}{@{}lcccc@{}}
\toprule
\multirow{1}{*}{Method}
 & \multicolumn{1}{c}{Success Rate$\uparrow$} 
 & \multicolumn{1}{c}{Success Rate Drop$\downarrow$} 
 & \multicolumn{1}{c}{x\_RMSE $\downarrow$} 
 & \multicolumn{1}{c}{y\_RMSE $\downarrow$}
 
 \\\midrule
\multirow{1}{*}{\textbf{DPPO\_adaptive}} 
& $\textbf{0.601}\pm0.418$ & $\textbf{38.42\%}$ & $0.272\pm0.063$           & $\textbf{0.097}\pm0.021$
\\ 
\midrule

\multirow{1}{*}{DPPO\_averse} ~\cite{schneider2024learning} 
& $0.327\pm0.268$      & $64.11\%$      & $\textbf{0.247}\pm0.025$  & $0.120\pm0.012$
\\ 
\midrule

\multirow{1}{*}{DPPO\_neutral} ~\cite{schneider2024learning} 
& $0.016\pm0.009$     & $98.32\% $     & $0.337\pm0.026$  & $0.144\pm0.004$ 
\\ 
\midrule

\multirow{1}{*}{PPO} 
& $0.029\pm0.003$      & $89.77\% $    & $0.300\pm0.001$  & $0.118\pm0.004$
\\ 
\midrule

\bottomrule
\end{tabular}%
}
\vspace{-15pt}
\end{table}

\subsection{Distribution Uncertainty Visualization}
\label{sec: vis}
\begin{figure} [htbp]
\centering
\includegraphics[width=0.49\textwidth, page=6, trim = 0.3cm 6.cm 5cm 0cm, clip]{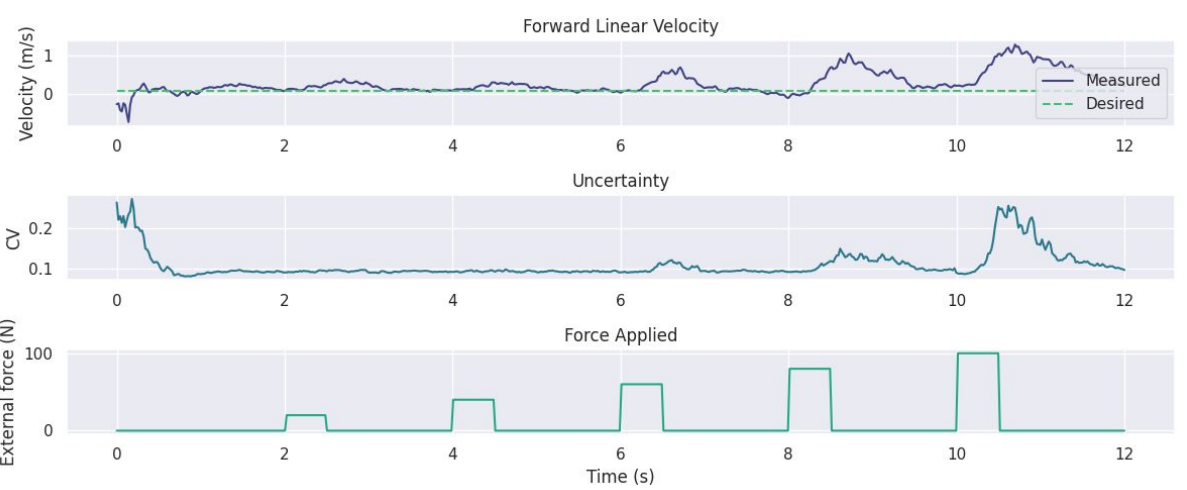} 
\caption{Uncertainties represented by the Coefficient of Variance (CV) visualized during evaluation with a 0 m/s target velocity, external forces are applied for 0.5s at 2s intervals.
}
\label{fig: risk}
\vspace{-10pt}
\end{figure}

To study how risk adaptiveness improves performance, we plot the uncertainty of the estimated return distribution, denoted by the coefficient of variation (CV) in Figure \ref{fig: risk}. 
The robot attempts to follow the 0m/s command, where external forces are applied for 0.5s at 2s intervals, with magnitudes ranging from 20N to 100N.
Each force exertion corresponds to an increase in both velocity deviation and uncertainty, showing temporary instability. 
After each peak, the uncertainty gradually declines, reflecting the policy effectively learns to regain stability.
With the coefficient of variation as a metric of the uncertainty of critic network distribution, we validate that the model identifies these higher-risk or uncertain situations caused by external perturbations.

At $t=0$, the uncertainty is high because the quadrupedal-to-bipedal transition involves less frequently visited states, resulting in greater uncertainty in the critic's value estimation. 
More optimistic actions can be taken in well-explored states, such as bipedal tracking without disturbances.
This underscores the importance of risk-adaptive DPPO in balancing conservatism in high-uncertainty states with optimism in well-explored states.
\begin{figure*} [htbp]
\centering
\includegraphics[width=1.0\textwidth, 
page=1, trim = 0.5cm 2.7cm 0cm 0cm, clip]
{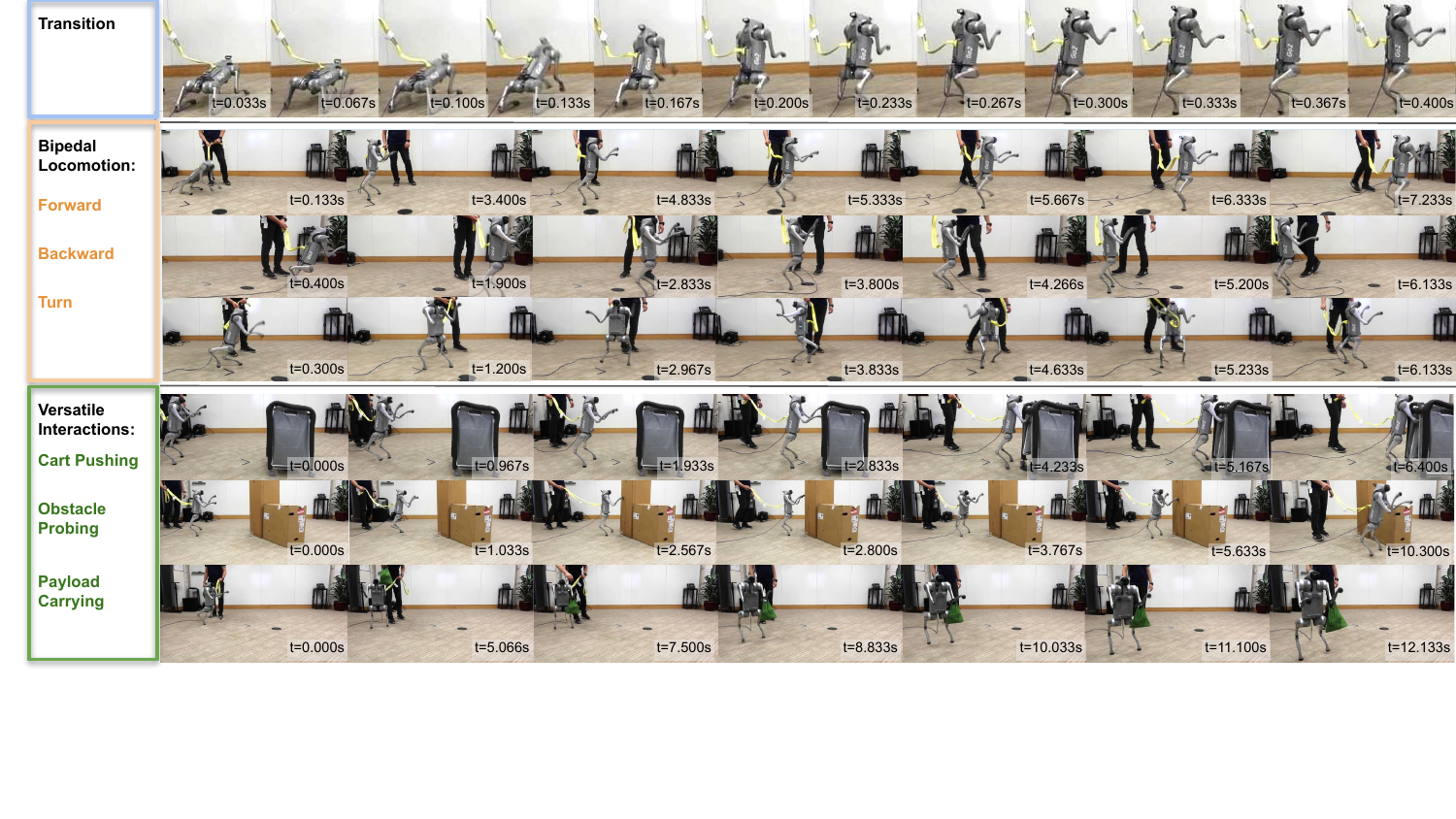}
\caption{Snapshots of bipedal loco-manipulation in the real world. 
From top to bottom, the images showcase the quadrupedal to bipedal transition, bipedal locomotion, and versatile interactions. 
Each row is marked with its respective timestamp for chronological analysis. 
Additional demonstrations, including velocity tracking at various speeds, clockwise and counterclockwise turns, and maintaining a stationary position, can be found in the supplementary video.}
\label{fig: real}
\vspace{-10pt}
\end{figure*}

\subsection{Real World Deployment}
\label{sec: real}
We deploy our policy on the Go2 robot in the real world, showcasing a single policy that enables versatile loco-manipulation capabilities, as illustrated in Figure \ref{fig: real}.
This policy enables not only basic locomotion such as forward, backward, and turning maneuvers but also supports complex interactions to further demonstrate its versatility. 
The robot can effectively fulfill tasks such as cart pushing, obstacle probing, and payload carrying. 
All of the real-world tasks introduce external forces that could destabilize the robot and require the robustness of the policy. 
Pushing a cart demands a robust loco-manipulation policy that can adjust the force applied and stabilize the body accordingly.
Obstacle probing requires the robot to recover from an unstable state when it runs into an obstacle and probes the obstacle with its front legs.
Carrying a payload increases the weight and shifts the robot's center of mass, necessitating dynamic balance and stability. 
Remarkably, the success of these real-world tasks is a direct result of the single bipedal locomotion policy, without requiring extensive task-specific training. This showcases the robustness and versatility of our bipedal locomotion policy and validates the effectiveness of our proposed risk-adaptive learning framework.

\section{Conclusion}
In this work, we introduce a risk-adaptive distributional RL framework for quadrupedal robots, enabling robust bipedal locomotion and versatile interactions with complex environments. 
Through extensive simulation and real-world experiments, we validate the robustness and adaptability of this framework, which is grounded in modeling the return distribution for risk-adaptive learning.
Refining the adaptation strategy for more flexible risk management could further enhance its performance. 
Future work may also focus on high-level planning and closed-loop control to facilitate long-horizon tasks.


\bibliographystyle{IEEEtran}
\bibliography{root}  

\end{document}